%% file: paper1_alt.tex
\documentclass[11pt]{article}

\usepackage[final]{acl}

\usepackage{times}
\usepackage{latexsym}
\usepackage[T1]{fontenc}
\usepackage[utf8]{inputenc}
\usepackage{microtype}
\usepackage{inconsolata}

\usepackage{graphicx}
\usepackage{float}
\usepackage{tikz}
\usepackage{amsmath,amssymb,amsfonts}
\usepackage{booktabs}
\usepackage{hyperref}
\hypersetup{pdftitle={Useful Diversity or Hallucinated Variety? Persona-Conditioned Synthetic Query Generation for E-Commerce Search},pdfauthor={Filippos Ventirozos, Tom Armitage, Mahmoud Oshagh, Peter Appleby, Matthew Shardlow}}
\usepackage{xcolor}
\usepackage{colortbl}
\usepackage{multirow}
\usepackage{enumitem}
\usetikzlibrary{arrows.meta,positioning}

\newcommand{\nllms}{6}
\newcommand{\nconfigs}{9}
\newcommand{\nsynth}{118{,}098}
\newcommand{\nhumaneval}{1{,}214}
\newcommand{\nprod}{15{,}304}
\newcommand{\nfields}{47}

\title{Useful Diversity or Hallucinated Variety?\\ Persona-Conditioned Synthetic Query Generation for E-Commerce Search}

\author{Filippos Ventirozos\textsuperscript{1,2}\quad Tom Armitage\textsuperscript{2}\quad Mahmoud Oshagh\textsuperscript{2} \\ \textbf{Peter A. Appleby\textsuperscript{2}}\quad \textbf{Matthew Shardlow\textsuperscript{1}} 
\\[2pt]
  \textsuperscript{1}Manchester Metropolitan University, Manchester, UK \\
  \textsuperscript{2}Autotrader UK, Manchester, UK \\
    \texttt{\{\href{mailto:f.ventirozos@mmu.ac.uk}{f.ventirozos}, \href{mailto:m.shardlow@mmu.ac.uk}{m.shardlow}\}@mmu.ac.uk}}

\begin{document}
\maketitle

\begin{abstract}
Synthetic supervision can bootstrap semantic parsers in new search domains, but prompt-induced diversity is useful only when generations are faithful to the structured intent. We explore this for cold-start e-commerce search: \nsynth{} synthetic pairs produced by \nllms{} LLMs with \nconfigs{} persona-conditioned prompt families on a \nfields{}-field UK automotive schema, against \nhumaneval{} human-annotated and \nprod{} production queries. We show that the IPIP personality bundle improves parser F1 in all four matched contrasts (Holm-corrected $p<.001$), and all three bundles have a positive main effect over the complete $2^3$ grid. Syntax conditioning is the axis that moves generations towards real traffic. It also induces lexical diversity. Alignment sKL, the divergence between von Mises--Fisher distributions fitted to synthetic and human-annotated queries, is an unlabelled proxy for parser F1 ($\rho=-0.917$ on intent-elicited human queries, lower is closer). An LLM committee, checked against a blinded author pass, triages the $927$ queries missed by every parser and labels $37\%$ as hallucinated. Dropped constraints dominate. The four syntax-bearing families have the largest hallucinated shares, $38$ to $42\%$, while IPIP alone has the lowest at $28\%$.
\end{abstract}

\section{Introduction}
\label{sec:introduction}
E-commerce platforms gain substantial UX and operational advantages by replacing faceted filters with natural-language interfaces. Form-based UIs scale poorly across vast catalogues~\citep{hearst2009search}, while modern shoppers embed complex constraints, such as long-tail specifications and compound Boolean logic, that static forms cannot accommodate~\citep{kelly2021starting,luo2023implicit, maarek2024boolean}. Recent approaches therefore frame e-commerce retrieval as semantic parsing, mapping raw NL queries directly into structured API fields~\citep{tagliabue2023vector,weld2022survey}. Training robust parsers requires extensive labelled data, making synthetic query generation an attractive scalable solution~\citep{amazon_cikm,2026_arizona}. The task is not open-ended dialogue but the narrower problem of mapping a structured search intent to a plausible natural-language query and training a smaller downstream model on those pairs. We study it in UK automotive search, a real, high-dimensional testbed with \nfields{} search fields and long-tail combinations of make, model, pricing, mileage, seller, finance, and feature constraints.

The central problem is not whether synthetic data can be made diverse. It mostly can. The problem is that useful diversity and schema faithfulness pull in different directions: aggressive prompting increases lexical and structural variation but can introduce unsupported attributes, flip inequality directions, or drift into assistant-style text that no longer matches the intent, while tightly constrained prompting preserves the schema and collapses outputs into repetitive templates. We use personas as an experimental mechanism for controlled variation along four axes (IPIP personality, attribute-style adjectives, syntactic modes, and UK demographic descriptors); their construction is in \S\ref{sec:prompt-types}.

Our main empirical contributions are threefold. First, every conditioned parser beats the vanilla prompt; the complete $2^3$ grid gives positive main effects for all three bundles, and the four matched contrasts put the IPIP bundle at $+1.1$ to $+2.1$ F1 points (\S\ref{sec:utility}). Second, we show that the alignment sKL between the synthetic corpus and a human reference is an unlabelled proxy for parser utility ($\rho=-0.917$ on our elicited reference). Additionally, we compare an in-house reference with \nprod{} live production queries (\S\ref{sec:alignment}). Third, a three-judge LLM committee audit of the $927$ universal parser failures, checked against a blinded author pass, locates the residual defects where syntax-bearing and demographic prompts drop the operator words that bind values to schema fields, and types them: dropped constraints dominate, unsupported additions and direction reversals follow (\S\ref{sec:judge-results}).

\section{Related Work}
\label{sec:related}
  \paragraph{Persona-conditioned synthetic generation.}
  LLM-generated supervision has become a standard substitute for human annotation \citep{wang2021want,honovich2023unnatural}, and persona prompting is now the
  dominant axis for injecting controllable variation into those generations. Persona Hub \citep{ge2024scaling} scales this to a billion personas, and a broader
  literature surveys role-play and persona-driven generation as a controllable-variation mechanism \citep{tseng2024twotales}. The mechanism rests on the
  observation that Big Five personality variables \citep{goldberg1992ipip} map onto measurable language cues \citep{mairesse2007personage}. Most of this work
  evaluates persona-conditioned outputs through surface diversity or downstream accuracy; \citet{kambhatla2025measuring} found that fine-grained persona detail adds
  little lexical diversity over a prompt-length control, which makes prompt length a live confound for any persona manipulation. Whether persona-induced
  variation remains faithful to a structured target intent is still open.

  \paragraph{Synthetic data generation for e-commerce search.}
  A parallel line of work applies LLM-generated supervision to e-commerce search. \citet{luo2023implicit} parse compound implicit constraints in Amazon product
  queries at scale, and \citet{amazon_cikm} build a natural-language product-search interface with synthetic supervision in a cold-start setting; more broadly,
  synthetic-query generation is now a standard tool for training retrievers and rankers without labels \citep{bonifacio2022inpars,chaudhary2023relative}. These papers evaluate synthetic queries through downstream retrieval or parser metrics, rarely by auditing the corpus itself against the structured catalogue fields the queries encode.

 \paragraph{Position.}
  The two closest prior works are by \citet{kambhatla2025measuring} on persona-induced lexical diversity in creative writing and by \citet{schaffelder2025synthetic} on source diversity in fine-tuning corpora. We share the question but instrument it against a structured e-commerce schema, with a three-lens audit per prompt configuration: internal diversity, distributional alignment to a held-out human reference and downstream parser utility, plus an LLM-judge audit of the residual failures.
   
\section{Synthetic Dataset}
\label{sec:syn-dataset}
We generated synthetic natural-language queries from a proprietary structured search dataset provided by our industry partner Autotrader UK, a large UK online automotive marketplace,\footnote{\url{https://www.autotrader.co.uk/}} amounting to 13M search intents over 7 days. Each example begins as a structured search intent: field--value constraints such as make, model, body type, mileage limit, price bound, or seller type. Several fields carry an \emph{operator} as well as a value, the relation that binds the value to the field: a price or mileage limit is a floor or a ceiling, a sort has a direction, and a location comes with a radius. In query text these relations are signalled by \emph{operator words} such as ``under'', ``within'', ``from'' and ``lowest''. We dampen high-frequency queries with a concave power transform to reduce duplicates. 


For each LLM, we use the same set of intents for all \nconfigs{} prompt settings. Between LLMs, however, the intent sets differ. The \nconfigs{} configurations are therefore directly comparable, since each one is generated from the exact same underlying intents, whereas between the LLMs we draw conclusions more broadly.


\subsection{Prompt Types and Motivation}
\label{sec:prompt-types}

We construct prompts from four different sources. Three of them (\texttt{p}, \texttt{a}, \texttt{s}) compose freely with one another, while the UK demographic axis \texttt{u} is used stand-alone, for reasons given below.

\paragraph{IPIP Big Five traits (\texttt{p}).}
To induce controlled linguistic variation, we steer register and phrasing using the IPIP Big Five personality axes \citep{goldberg1992ipip}: openness modulates exploratory versus conventional wording, conscientiousness governs explicit constraint marking, extraversion affects enthusiasm, agreeableness influences politeness, and neuroticism controls tentativeness. Prior work shows that these signals are reliably interpretable, with \citet{jiang2024personallm} reporting approximately 80\% rater accuracy for intended traits, and that they remain stable across frontier models \citep{sorokovikova2024big5}, which is important for multi-generator comparability. Each axis is sampled at three levels (high, low, neutral), with probability $0.3$ assigned to the neutral state, yielding up to $3^5$ combinatorial profiles per generation call (Appendix~\ref{app:ipip}).

\paragraph{UK demographic descriptors (\texttt{u}).}
Motivated by NVIDIA's US-demographic persona set \citep{Nemotron-Personas-USA} and by evidence that structured demographic profiles aid realistic synthetic user conversations \citep{yun2025sleepless}, we build UK demographic profiles from a UK Open Data consumer-segmentation source. Each frequent postcode area carries 57--221 attribute flags (median 132) spanning age, income, household, and lifestyle; per call we sample a postcode and draw 3--15 of its attributes (Appendix~\ref{app:uk-persona}).

\paragraph{Syntax modes (\texttt{s}).}
Following the Yahoo! web-query-log grammar of \citet{barr2008linguistic}, in which web queries form a grammar of their own dominated by noun phrases, we add an explicit syntax axis with six modes (noun phrases, telegraphic fragments, natural-language questions, imperatives, post-modified noun phrases, navigational fragments), weighted as in that study and each accompanied by two sampled in-prompt examples (Appendix~\ref{app:syntax}).

\paragraph{Attribute-style adjectives (\texttt{a}).}
Shoppers in our partner's beta tests described cars with vague adjectives such as \emph{reliable}, \emph{cheap} and \emph{decent}, which the schema has no field to hold. This is the vocabulary problem in human-system communication \citep{furnas1987vocabulary}: people name a thing differently from the system that stores it. We therefore add an axis that supplies such words deliberately, together with guards on when each one fits. It draws on eight dimensions: \texttt{hard\_specs}, \texttt{aesthetic} (\emph{sleek}), \texttt{kinetic} (\emph{fast}), \texttt{affordance} (\emph{family}), \texttt{evaluative} (\emph{pristine}), \texttt{economic} (\emph{affordable}), \texttt{social} (\emph{luxury}) and \texttt{vague}. One dimension is sampled per call, and a semantic-fit guard blocks mismatches. For instance, prestige wording is reserved for luxury models (Appendix~\ref{app:adjectives}).

\paragraph{Prompt families.}
Table~\ref{tab:prompt-families} in Appendix~\ref{app:prompts} lists the \nconfigs{} prompt families. Here \texttt{n} is the vanilla prompt: the model receives only the structured intent and a generic instruction to produce a faithful shopper query. The three text-style bundles (\texttt{p}, \texttt{a}, \texttt{s}) compose freely, yielding the complete $2^3$ grid: \texttt{n}, three single-bundle families, three pairwise families and the full \texttt{pas}. The UK demographic bundle stays stand-alone (\texttt{u}). A single postcode draw already injects 3--15 attributes from a pool of 57--221 flags, saturating the prompt with multi-cue conditioning, so mixing it with \texttt{p}, \texttt{a} and \texttt{s} would overflow the prompt budget and push us into the four-axis regime where mixing complete personas has been reported to degrade alignment \citep{bui2025mixturepersonas}. We therefore read \texttt{u} as an exploratory comparator rather than a fourth axis. 


\paragraph{Generation pool.}
We chose \nllms{} state-of-the-art reasoning-mode LLMs, spanning closed-source frontier APIs (GPT-5.4 \citep{openai2026gpt54}, Claude Opus 4.6 \citep{anthropic2026opus46}, Gemini 3.1 Pro \citep{deepmind2026gemini31pro}) and open-weight reasoning models (DeepSeek-R1-Distill-Llama-70B \citep{deepseek2025r1}, Hunyuan-A13B-Instruct \citep{tencent2025hunyuana13b}, Qwen3-Next-80B-A3B-Thinking \citep{qwen2025next}). We held reasoning mode constant across the pool because chain-of-thought generation supports complex reasoning \citep{wei2022chain}, which we expect to translate into schema faithfulness. Under the per-LLM intent-paired design, each LLM contributes $\nsynth{}/\nllms{}$ queries and each configuration's slice holds $\nsynth{}/\nconfigs{}$ pooled across generators (Appendix~\ref{app:generators}).
\subsection{Diversity}
\label{sec:diversity}

We report two measures of internal lexical diversity. \textbf{Distinct-2} \citep{li2016diversity} rewards fresh bigrams, so a higher score means a more diverse corpus. \textbf{Self-BLEU-4} \citep{zhu2018texygen} goes the other way. It measures how much a corpus repeats itself, so lower values indicate greater diversity. Because the intents are matched across prompt configurations within each generator (\S\ref{sec:prompt-types}), we compute both metrics separately for every (LLM, configuration) cell. Each cell is scored on its own train and validation pool, meaning everything in that cell's corpus except for the reserved synthetic test set (\S\ref{sec:parser-method}). For each configuration, we then take the mean and standard deviation across LLMs.

\begin{figure}[t]
\centering
\includegraphics[width=\linewidth]{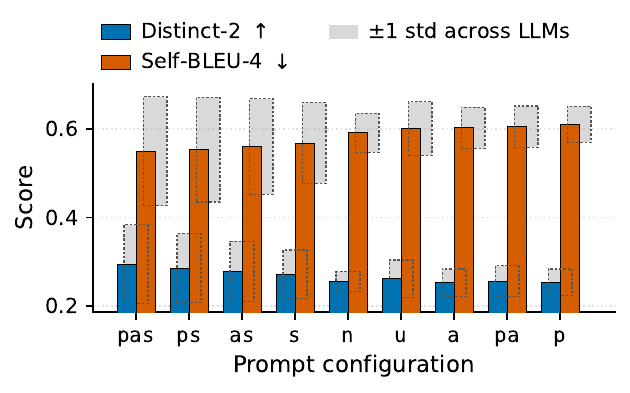}
\caption{Internal diversity per prompt configuration, ordered by Self-BLEU-4 (more diverse on the left). Coloured bars are the mean across \nllms{} generators; the light-grey shadow behind each bar spans $\pm 1$ std across LLMs. Tall shadows flag configurations whose diversity depends heavily on the generator; short shadows indicate that the prompt itself, not the LLM, is doing the work.}
\label{fig:internal-diversity}
\end{figure}

Figure~\ref{fig:internal-diversity} orders the \nconfigs{} configurations by Self-BLEU-4 (the more reliable of the two on a fixed pool). Two patterns dominate. First, syntax-bearing families are the diversity engine: \texttt{ps}, \texttt{pas}, \texttt{as}, and \texttt{s} lead on both metrics. Second, the cross-LLM spread itself grows with syntax: the grey shadows on \texttt{ps} and \texttt{pas} are roughly three times wider than on \texttt{n}, meaning a syntax instruction lets generator identity express itself --- Gemini-3.1 and Opus-4.6 lean into the new register and lift diversity sharply, while three stay close to their default phrasing and Qwen3 becomes more repetitive (Table~\ref{tab:per-llm-selfbleu}). Syntax-free conditioning is mostly a lexical nudge with tight cross-LLM agreement: \texttt{p}, \texttt{a}, and \texttt{pa} sit within 0.003 Distinct-2 of \texttt{n}. The demographics-only family \texttt{u} is the only syntax-free prompt with a Distinct-2 lead over \texttt{n}, and a small one.

\section{Human Dataset}
\label{sec:human-dataset}

We collected \nhumaneval{} \texttt{(NL query, structured intent)} pairs through a University annotation campaign. The campaign ran two protocols. In the first, annotators write a free-form query and then express it by hand in the partner's filter interface (\S\ref{sec:human-at}). Under the second, annotators see a fixed, structured intent on a custom HTML page and write a query for it in their own words (\S\ref{sec:human-html}). Ecological validity is higher in the first. The second protocol gives some of that up, in return for coverage breadth and for the intent matching that the alignment analysis needs. It also supplies $87\%$ of the corpus. Because the intents are fixed in advance, the corpus spans vehicle types, price tiers and buying scenarios beyond the subset the recruited annotators would gravitate towards on their own.

Both protocols draw on the same pool of nine annotators: seven took part in the persona-card protocol and six in the website-front verification, with four in both. All were screened to give a mix of age, gender and self-rated car knowledge via a pre-task questionnaire. Self-rated car knowledge ranged from 1 to 4 on a 5-point scale, so the pool holds novices, intermediates and enthusiasts side by side. Five of the nine were planning to buy a car, either actively or in the near term, and prior familiarity with the partner platform ranged from frequent users to people who had never opened it. The spread is deliberate, so the corpus exhibits between-user variation rather than a single domain-expert style. Appendix~\ref{app:rise-iaa} gives the agreement metrics and filtering.

\subsection{Protocol 1: Verification through the Partner's Website Front}
\label{sec:human-at}

Annotators write a free-form NL query and then express it themselves on the partner's ordinary filter-based search page, choosing make, model, body type, price, mileage and the rest by hand and recording the resulting URL. The structured intent is therefore the annotator's own reading of their query rather than a system's parse, and these are the only rows written without a schema or persona stimulus in view. A second independent pass over the same items supplies double coverage. The protocol yielded 156 pairs after filtering (Appendix~\ref{sec:appendix:filtering}), across six 30-item batches.
\subsection{Protocol 2: Custom HTML Page Experiment}
\label{sec:human-html}

Annotators see each structured intent as a third-person scenario card and type their own NL query into a browser form. GPT-5.2 wrote the cards. We settled on it after comparing how several LLMs structure such scenarios. A field list beside the card---the same field--value list the card was written from---lets annotators verify that their query covers all intended constraints. Below, we refer to this as the \emph{persona-card} protocol. It yielded 1{,}058 retained pairs after filtering (Appendix~\ref{app:rise-iaa}).


\subsection{External Reference: Live Production Traffic}
\label{sec:production-reference}

To obtain a reference that no annotation stimulus shaped, we additionally take one month of natural-language queries submitted to the partner's deployed search interface. We never score a parser on it: its query text feeds the alignment analysis, and its recorded fields enter only the corpus comparison of \S\ref{sec:data-stats}. Every logged string was screened for personal data and matching rows discarded (Ethics Statement), leaving \nprod{} of the $15{,}562$ logged queries.

\subsection{Data Statistics}
\label{sec:data-stats}
The synthetic corpus contains \nsynth{} \texttt{(query, intent)} pairs across the \nconfigs{} prompt configurations; the held-out human reference contains \nhumaneval{} pairs (1{,}058 persona-card, 156 website-front); the unlabelled production reference contains \nprod{} queries. Synthetic examples average 5.96 structured fields and 17.6-word queries, human examples 3.17 fields and 14.3 words, and production queries 4.8 words (median 3). Real queries are short because most are lookups rather than constrained requests: $19.2\%$ are a single word, $83.0\%$ contain no operator word at all, and among queries of two words or fewer that the deployed engine parsed, $62.4\%$ resolve to a make or a model and nothing else. \emph{Mitsubishi outlander}, \emph{BMW 330i}, \emph{electric corsa}, \emph{wheelchair accessible} and \emph{LHD} (left-hand drive) are typical. The length gap is real (Table~\ref{tab:ecological-gap}): $47.3\%$ of production queries are two words or shorter, against $0.4\%$ of the human reference and $0.9\%$ of the synthetic corpus. Field coverage on both annotated sides follows the marketplace distribution, with \texttt{search\_make} and \texttt{search\_model} most frequent.

\subsection{Alignment to Human and Production References}
\label{sec:alignment}
\label{sec:production-alignment}


We fit von Mises--Fisher distributions \citep{banerjee2005vmf} to each synthetic and each reference embedding cloud (OpenAI \texttt{text-embedding-3-large}, unit-normalised) by maximum likelihood with the Banerjee--Sra high-dimensional approximation. We report \textbf{Spread $\kappa$} (the synthetic-side concentration; lower is broader) and \textbf{Alignment sKL} (the symmetric KL divergence between a synthetic fit and a reference fit; lower is closer). Repeating the measurement against the production corpus of \S\ref{sec:production-reference}, with parser F1 (\S\ref{sec:parser-method}) held at its in-house value, isolates one question: whether the ranking depends on how the reference was elicited (Table~\ref{tab:reference-alignment-matrix}).

\input{arr_revision/reference_alignment_matrix.tex}


The two configurations that pair personality with syntax, \texttt{pas} and \texttt{ps}, produce the synthetic queries that are closest to every reference, including real traffic. The adjectives-only configuration \texttt{a} is always the furthest or second furthest. Its queries and those of the vanilla prompt \texttt{n} form the two most concentrated clouds ($\kappa$). The two references written without a stimulus in view, website-front (protocol 1) and real traffic, order all nine configurations almost identically (Kendall $\tau=0.944$). Both put the four syntax-bearing configurations closest, because \texttt{s} shortens queries and strips grammatical glue (Appendix~\ref{app:surface}).

Persona-card (protocol 2) queries have a different register. Annotators write them in response to a descriptive scenario card, and the queries are longer and mostly in the first person (15.0 words on average, compared with 9.5 for website-front). An annotation stimulus shapes the data collected under it \citep{parmar2023dont}. That first-person register is what \texttt{p} and \texttt{u} add to generated queries (Appendix~\ref{app:surface}), so the persona-card reference moves \texttt{u} and \texttt{pa} up the ranking and \texttt{s} down ($\tau=0.500$ against real traffic).

The parsers are scored mostly on persona-card rows, yet the IPIP gains in F1 also hold on the free-form website-front queries, in three of the four contrasts (Table~\ref{tab:protocol-contrasts}). Alignment to the persona-card reference tracks parser F1 closely ($\rho=-0.917$, $p<.001$). The correlation stays strong against the full in-house (protocols 1 \& 2) reference ($\rho=-0.800$, Table~\ref{tab:alignment-bootstrap}), but against real traffic it is weak ($\rho=-0.350$, $p=.36$). Each reference answers its own question. Label-free alignment sKL therefore tracks parser utility when the reference matches the evaluation queries, and measures realism when the reference is real traffic.

\section{Methodology}
\label{sec:methodology}

\subsection{Per-Configuration Parser}
\label{sec:parser-method}

We fine-tuned a Flan-T5-Large parser \citep{chung2024flan} for each prompt configuration on 10{,}910 instances of that configuration alone, with 1{,}212 for validation, plus a \emph{mixed-corpus} parser on a training set of the same size sampled uniformly across all \nconfigs{} configurations. All $\nconfigs{}+1$ parsers were fully fine-tuned with best-checkpoint selection on validation F1 (training details in Appendix~\ref{app:robustness}) and were evaluated with greedy decoding on the \nhumaneval{}-pair held-out human reference (\S\ref{sec:human-dataset}) and a reserved synthetic test set. Following the tuple-level F1 used in aspect-based sentiment analysis (ABSA) \citep{pontiki2016semeval}, a predicted \texttt{(field, value)} tuple counts as a true positive only if both the field name and its value match the gold, regardless of prediction order.

\subsection{LLM-as-Judge Committee}
\label{sec:judge-method}

To triage T5 errors into likely parser-side failures and likely unfaithful synthetic instances, we use a three-judge LLM committee. Each judge is a current frontier reasoning model from one of the closed-source generator families: \texttt{claude-opus-4-7}, \texttt{gpt-5.5-pro} and \texttt{gemini-3.1-pro-preview}, each invoked at its highest reasoning setting. Per-instance labels are taken by majority vote. The committee was asked one binary question, whether the generator hallucinated the intent, under a \emph{cleanroom} prompt exposing only the seed \texttt{(field, value)} tuples and the resulting query, with no LLM identity, prompt-configuration code, or parser predictions visible. That rubric is more permissive than the metric the parser is scored on, accepting ``Japanese car'' for \texttt{make}${=}$\texttt{Honda} and ``low mileage'' for an exact $40{,}000$-mile bound, which the exact field--value metric counts as errors. Its labels are therefore a triage signal for locating defects, not faithfulness scores on the parser's scale. Full prompts, model settings, aggregation, reliability, and one scoring rule for the no-limit search radius are in Appendix~\ref{app:judge}.

\section{Results}
\label{sec:results}
We report parser utility per configuration (\S\ref{sec:utility}) and a conditional universal-fail audit by configuration and generator (\S\ref{sec:judge-results}).

\subsection{Parser Utility on Human Eval}
\label{sec:utility}

Table~\ref{tab:utility-perconfig} reports F1 with field--value precision and recall on the held-out \nhumaneval{}-pair human reference for the $\nconfigs{}+1$ per-configuration T5 parsers.

\begin{table}[t]
\centering
\small
\setlength{\tabcolsep}{3pt}
\begin{tabular}{lccc}
\toprule
\textbf{Cfg} & \textbf{F1} & \textbf{Prec.} & \textbf{Rec.} \\
\midrule
\texttt{n}     & 75.98 $\pm$ 0.22 & 73.57 $\pm$ 0.40 & 80.94 $\pm$ 0.14 \\
\texttt{s}     & 76.91 $\pm$ 0.20 & 74.67 $\pm$ 0.24 & 81.71 $\pm$ 0.29 \\
\texttt{a}     & 77.70 $\pm$ 0.20 & 76.55 $\pm$ 0.33 & 80.87 $\pm$ 0.18 \\
\texttt{p}     & 77.99 $\pm$ 0.23 & 76.40 $\pm$ 0.39 & 81.69 $\pm$ 0.07 \\
\texttt{as}    & 78.55 $\pm$ 0.27 & 77.07 $\pm$ 0.17 & 82.21 $\pm$ 0.45 \\
\texttt{ps}    & 79.05 $\pm$ 0.40 & 77.74 $\pm$ 0.57 & \textbf{82.50} $\pm$ 0.11 \\
\texttt{u}     & 79.19 $\pm$ 0.55 & \textbf{79.10} $\pm$ 0.78 & 81.36 $\pm$ 0.42 \\
\texttt{pa}    & 79.40 $\pm$ 0.31 & 78.85 $\pm$ 0.36 & 81.91 $\pm$ 0.25 \\
\texttt{pas}   & \textbf{79.62} $\pm$ 0.40 & 78.71 $\pm$ 0.47 & 82.47 $\pm$ 0.34 \\
\midrule
\texttt{mixed} & 79.36 $\pm$ 0.18 & 78.71 $\pm$ 0.37 & 82.02 $\pm$ 0.07 \\
\bottomrule
\end{tabular}
\caption{Per-configuration parser utility on the held-out human reference, tuple-level F1 with field--value precision and recall, mean $\pm$ std across three train/val resamples (test set held fixed, all numbers $\times 100$). \texttt{mixed} is the uniform-sample baseline pooling all \nconfigs{} configurations at matched training volume.}
\label{tab:utility-perconfig}
\end{table}

The vanilla baseline \texttt{n} underperforms every conditioned configuration, by $0.9$ to $3.6$ F1 points. We test the four planned IPIP contrasts with a paired bootstrap over the \nhumaneval{} human test queries ($100{,}000$ resamples, Appendix~\ref{app:robustness}). The IPIP bundle improves F1 in all four contrasts. The gains are $+2.01$ points for \texttt{n}$\rightarrow$\texttt{p}, $+1.70$ for \texttt{a}$\rightarrow$\texttt{pa}, $+2.14$ for \texttt{s}$\rightarrow$\texttt{ps} and $+1.08$ for \texttt{as}$\rightarrow$\texttt{pas}. All four $95\%$ confidence intervals exclude zero. Holm-corrected $p$-values are below $.001$ throughout. These gains are attributable to the whole \texttt{p} bundle (\S\ref{sec:prompt-types}), and the effect of personality alone cannot be isolated from it. Recall exceeds precision in every configuration (80.87--82.50 against 73.57--79.10), so the parsers recover the intended fields but also produce false-positive tuples.

The bundles \texttt{p}, \texttt{a} and \texttt{s} form a complete $2^3$ grid. The same rows, therefore, give main effects and interactions for all three bundles. Averaged over the other two bundles, the main effects are $+1.73$ F1 points for \texttt{p} ($95\%$ confidence interval $1.40$ to $2.07$), $+1.34$ for \texttt{a} ($1.05$ to $1.63$) and $+0.76$ for \texttt{s} ($0.42$ to $1.11$). All three intervals exclude zero. Combining bundles still helps, but their gains overlap. A pairwise interaction measures how much smaller one bundle's gain is when the other is already present. Every interaction is negative. The values are $-0.68$ F1 points for \texttt{pa}, $-0.46$ for \texttt{as}, $-0.25$ for \texttt{ps} and $-0.75$ for the three-way \texttt{pas} term. Adding \texttt{p}, for instance, is worth $+2.01$ F1 on the vanilla prompt and $+1.08$ on top of \texttt{as} (\texttt{n}$\rightarrow$\texttt{p} against \texttt{as}$\rightarrow$\texttt{pas}). The \texttt{pas} parser scores highest, but only by $0.22$ F1 points over \texttt{pa} and $0.27$ over \texttt{mixed}. The confidence intervals of both differences include zero. The three parsers are therefore statistically indistinguishable.

The IPIP gains hold when we split the human reference by collection protocol (persona-card or website-front) or by the number of fields in the gold intent (0--2, 3, or 4+). Across these five subsets, all $20$ gains (four contrasts per subset) stay positive, and $16$ of their $20$ confidence intervals exclude zero (Appendix~\ref{app:robustness}, Table~\ref{tab:protocol-contrasts}).

\subsection{Conditional Universal-Fail Audit}
\label{sec:judge-results}
We next audit the synthetic test \emph{mixed-corpus}, a $9{,}000$-row pool uniformly sampled across the \nconfigs{} prompt configurations. A query missed by every parser, across every configuration and all three resamples, is unlikely to be one run's idiosyncrasy, and it isolates the generator's central question: is the query a faithful but rare expression of the intent, or did generation drift from the intended field--value set? We mark a query \emph{universally failed} if all $3 \times (\nconfigs{}+1)=30$ runs fail to parse it perfectly, yielding $927$ rows, about $10\%$ of the pool. The estimand is explicitly conditional: we measure $P(\text{hallucination}\mid\text{universal failure})$, which is not $P(\text{hallucination})$.

\paragraph{Where the hard cases sit.}
Hard cases concentrate in the demographic and syntax-bearing prompts (Table~\ref{tab:gen-x-config-hallucination}, denominators): \texttt{u} contributes the most rows ($129$) and \texttt{n} and \texttt{a} the fewest ($73$ and $72$). Column totals are comparable because, within each generator, the same intent indices are verbalised under every configuration (\S\ref{sec:syn-dataset}), so no configuration is favoured by an easier slice of intents. Row totals are not: each generator holds an equal share of the test pool, but each LLM verbalised its own intent index set, so Opus-4.6's $415$ rows against GPT-5.4's $62$ count universal failures on different intents rather than ranking the generators.

\paragraph{Hallucination audit.}

Within the $927$ universal failures, the committee labels $345$ rows ($37.2\%$) as hallucinated and $582$ ($62.8\%$) as faithful (Table~\ref{tab:gen-x-config-hallucination}). \emph{By prompt configuration}, the share of hallucinated rows rises from $28.2\%$ to $42.3\%$ in the order $\texttt{p} < \texttt{n} < \texttt{a} < \texttt{pa} < \texttt{u} < \texttt{s} < \texttt{as} < \texttt{ps} < \texttt{pas}$. Every judge places \texttt{p}, \texttt{n} and \texttt{a} lowest. The three multi-axis syntax configurations, \texttt{as}, \texttt{ps} and \texttt{pas}, sit in the upper half of every judge's ordering, while the positions of \texttt{s}, \texttt{u} and \texttt{pa} vary by judge. \emph{By generator}, Opus-4.6 ($124$ rows) and Gemini-3.1 ($74$) together account for $57\%$ of the $345$ hallucinated rows. For both generators, about two thirds of their hallucinated rows fall in the four syntax-bearing configurations ($63\%$ and $70\%$). Qwen3 is a clean outlier ($0/78$). Hunyuan-A13B and DeepSeek have the highest hallucinated share of their own rows ($63/101$ and $65/116$). Read together, the two orderings locate the hallucinated universal failures. Among the queries that every parser missed, those written under a syntax-bearing configuration are the most likely to have drifted from their intent ($42.3\%$ for \texttt{pas} against $28.2\%$ for \texttt{p}). Opus-4.6 and Gemini-3.1 produce the most drifted queries, largely under those same configurations. Hunyuan-A13B and DeepSeek produce fewer, but a larger fraction of their own universal failures has drifted.

\begin{table*}[t]
\centering
\small
\setlength{\tabcolsep}{4pt}
\begin{tabular}{l|ccccccccc|c}
\toprule
\textbf{LLM} & \texttt{n} & \texttt{u} & \texttt{p} & \texttt{a} & \texttt{s} & \texttt{pa} & \texttt{ps} & \texttt{as} & \texttt{pas} & \textbf{All} \\
\midrule
DeepSeek   & \cellcolor{red!4}2/9 & \cellcolor{red!18}8/13 & \cellcolor{red!16}7/13 & \cellcolor{red!16}7/13 & \cellcolor{red!9}4/10 & \cellcolor{red!24}11/17 & \cellcolor{red!18}8/13 & \cellcolor{red!20}9/13 & \cellcolor{red!20}9/15 & 65/116 \\
Gemini-3.1 & \cellcolor{red!7}3/9 & \cellcolor{red!31}14/20 & \cellcolor{red!4}2/6 & \cellcolor{red!4}2/10 & \cellcolor{red!31}14/28 & \cellcolor{red!2}1/7 & \cellcolor{red!18}8/17 & \cellcolor{red!29}13/31 & \cellcolor{red!38}17/27 & 74/155 \\
GPT-5.4    & \cellcolor{red!4}2/7 & \cellcolor{red!4}2/6 & \cellcolor{red!7}3/8 & \cellcolor{red!4}2/5 & \cellcolor{red!4}2/7 & \cellcolor{red!4}2/6 & \cellcolor{red!4}2/8 & \cellcolor{red!4}2/6 & \cellcolor{red!4}2/9 & 19/62 \\
Hunyuan    & \cellcolor{red!22}10/16 & \cellcolor{red!22}10/14 & \cellcolor{red!7}3/7 & \cellcolor{red!11}5/8 & \cellcolor{red!29}13/16 & \cellcolor{red!16}7/11 & \cellcolor{red!11}5/10 & \cellcolor{red!9}4/9 & \cellcolor{red!13}6/10 & 63/101 \\
Opus-4.6   & \cellcolor{red!13}6/20 & \cellcolor{red!31}14/65 & \cellcolor{red!20}9/43 & \cellcolor{red!18}8/29 & \cellcolor{red!33}15/57 & \cellcolor{red!20}9/33 & \cellcolor{red!60}\textbf{27/64} & \cellcolor{red!40}18/50 & \cellcolor{red!40}18/54 & \textbf{124/415} \\
Qwen3      & 0/12 & 0/11 & 0/8 & 0/7 & 0/9 & 0/8 & 0/8 & 0/7 & 0/8 & 0/78 \\
\midrule
\textbf{All} & \cellcolor{red!27}23/73 & \cellcolor{red!55}48/129 & \cellcolor{red!28}24/85 & \cellcolor{red!28}24/72 & \cellcolor{red!55}48/127 & \cellcolor{red!35}30/82 & \cellcolor{red!58}50/120 & \cellcolor{red!53}46/116 & \cellcolor{red!60}\textbf{52/123} & \textbf{345/927} \\
\bottomrule
\end{tabular}
\caption{Committee majority-vote labels within the conditional universal-fail subset, under the rubric of \S\ref{sec:judge-method}. Each cell shows $n_\text{hallu}/n_\text{majority}$ among the $927$ universal failures after the scoring rule of Appendix~\ref{app:judge}, shaded by absolute $n_\text{hallu}$. Configurations enter this pool at different rates, so these composition counts estimate neither configuration-level nor corpus-level hallucination rates.}
\label{tab:gen-x-config-hallucination}
\end{table*}

\paragraph{Manual author audit.}
On a random sample of $100$ cases, drawn in proportion to configuration, a two-pass author audit (a blinded first pass followed by adjudication of the disagreements) agrees with the committee majority on $83\%$ of cases (Appendix~\ref{app:judge}). The author labels $25$ cases as hallucinated: nine omitted constraints (most often the sort order), six omitted search radii of $40$ to $100$ miles, nine altered values (five of which reverse a direction) and one ambiguous numeric case. The committee flags more than the author accepts, with precision $0.60$ at recall $0.96$ against the author's labels, which is why we read it as a triage instrument (\S\ref{sec:judge-method}).

\paragraph{Hallucination types.}

Each judge also gives a short rationale for its verdict. We tag the rationales of all $948$ hallucinated verdicts, up to three per query, with deterministic cue lexicons for four non-exclusive patterns (Appendix~\ref{app:judge}). The resulting shares are cue rates, not a partition of errors. \textbf{Dropped constraints} dominate ($57.3\%$), especially under \texttt{u} and the syntax-bearing prompts. \textbf{Unsupported additions} are less common ($15.7\%$), and their commonest form is quietly widening a discrete field: an intent for five doors gets written as ``3 or 5 doors''. Under the \texttt{u} configuration a second kind shows up, with Gemini-3.1 carrying the sampled persona's town and price preference into the query. An intent with no location and no price becomes ``cheapest Volvo XC40 SUV 2018 or newer within 30 miles of Doncaster''. \textbf{Direction reversals} are rarer ($6.8\%$) and concentrated in Hunyuan-A13B and DeepSeek. They sit on ordered fields: a \emph{minimum} price of \pounds500 becomes ``under \pounds500'' and a descending price sort becomes ``lowest to highest''. \textbf{Swapped values} are essentially absent ($0.2\%$). The two flagged rationales come down to one intent, where the ``Petrol Hybrid'' fuel type was written as plain ``Petrol''.

The errors under syntax prompts seem to come from a function-word tax: noun-phrase and telegraphic modes drop words such as ``under'', ``within'', ``from'', and ``lowest'', which often carry the schema operator. Those two modes have the largest hallucinated shares, $43.4\%$ and $41.8\%$, while post-modified noun phrases, which keep the relational prepositions, carry the smallest, $25.7\%$ (Table~\ref{tab:syntax-hallucination}). Syntax prompting still helps, but it should not blindly strip prepositions or stop-words.

\section{Discussion}
\label{sec:discussion}

We read each axis through the four lenses of the paper: utility is parser F1 (\S\ref{sec:utility}), realism is alignment sKL to real traffic (\S\ref{sec:alignment}), diversity is Distinct-2 and Self-BLEU-4 (\S\ref{sec:diversity}), and faithfulness is the audit (\S\ref{sec:judge-results}).

\paragraph{Syntax (\texttt{s}).}
\textbf{Realism:} the only axis that moves generations towards real traffic in every matched contrast ($-60.7$ to $-116.1$ sKL, Table~\ref{tab:axis-skl-contrasts}). It works by deleting grammatical glue. Adding \texttt{s} cuts the function-word share by $2.5$ points and the share of queries carrying an operator word by $4.2$ points (Appendix~\ref{app:surface}), and real traffic is $20.8\%$ function words against $42.3\%$ in our reference. \textbf{Diversity:} the four syntax-bearing families lead on both metrics, and the cross-LLM spread grows with syntax (\S\ref{sec:diversity}). \textbf{Utility:} the smallest main effect of the three, $+0.76$ F1. \textbf{Faithfulness:} the deleted glue includes the operator words (``under'', ``within'', ``from''). Their absence is the commonest cue in the hallucinated rationales, and the four syntax-bearing configurations have the highest hallucinated shares in the audited pool (\S\ref{sec:judge-results}). Realism and faithfulness risk may therefore be two sides of one operation.

\paragraph{Personality (\texttt{p}).}
\textbf{Utility:} the most reliable gain of the three bundles, $+1.73$ F1 as a main effect and positive in all four matched contrasts (\S\ref{sec:utility}). \textbf{Realism:} the gain does not come from moving closer to real traffic, most of which is short keyword queries. The bundle makes queries longer ($+3.4$ words) and more often written in the first person, ``I'' or ``my'' ($+7$ points, Appendix~\ref{app:surface}): on the vanilla prompt it moves generations away from real traffic ($+24.9$ sKL), while towards the reference our annotators wrote it helps in every contrast ($-17.3$ to $-37.6$ sKL, Table~\ref{tab:axis-skl-contrasts}). A persona prompt does not turn the model into that person \citep{zheng2024helpful,hu2024quantifying}. \texttt{p} works instead like paraphrasing, which widens wording but leaves sentence structure intact \citep{niu2023bridging}: operator words and intent stay, and the parser trains on more wordings of each field--value set. \textbf{Diversity:} a lexical nudge only (\S\ref{sec:diversity}). \textbf{Faithfulness:} \texttt{p} is low on hallucinations ($28.2\%$ of its hard cases, the lowest share in Table~\ref{tab:gen-x-config-hallucination}), which makes it the cleanest case of useful diversity. The tension between diversity and label correctness reported for LLM-generated data \citep{chung2023increasing} is therefore two dials rather than one, and our axes turn them separately. The lexical dial, \texttt{p}, buys utility. The form dial, \texttt{s}, buys realism at the plausible cost of operator words, and \texttt{pas} turns both.

\paragraph{Adjectives (\texttt{a}).}
\textbf{Utility:} \texttt{a} barely changes query length or word choice (Appendix~\ref{app:surface}), yet its gain is almost all precision, the vanilla parser's weak side (precision $73.6$ against recall $80.9$): $+2.20$ precision, $+0.15$ recall, $+1.34$ F1 (main effects derived from Table~\ref{tab:utility-perconfig}). \textbf{Realism:} alone, \texttt{a} is the furthest configuration from real traffic ($+42.9$ sKL over \texttt{n}), but under \texttt{s} it helps ($-10.6$ and $-9.0$). \textbf{Diversity:} none beyond \texttt{n}. The adjectives are given as suggestions in the prompt when they match semantically (Appendix~\ref{app:adjectives}). Mostly Gemini-3.1 and Opus-4.6 use them, and only their cells move on Distinct-2. \textbf{Faithfulness:} $33\%$ of its hard cases are hallucinated, below the pooled $37\%$ (Table~\ref{tab:gen-x-config-hallucination}).

\paragraph{Demographics (\texttt{u}).}
We did not cross \texttt{u} with the other axes, so it has no matched contrasts, and the UK-exposure caveat of \S\ref{sec:limitations} applies. \textbf{Utility:} the highest precision of the nine parsers ($79.10$) and third on F1. It also has the widest coverage of frequent production vocabulary, $73.6\%$ of the thousand most frequent content words against $60.9\%$ for \texttt{n} (Appendix~\ref{app:surface}). \textbf{Realism:} real traffic ranks it fifth of nine, behind every syntax-bearing configuration. The in-house reference ranks it third (\S\ref{sec:alignment}). \textbf{Diversity:} a small Distinct-2 lead over the other syntax-free prompts ($+0.007$ over \texttt{n}) and no gain on Self-BLEU-4. \textbf{Faithfulness:} $37\%$ of its hard cases are hallucinated, equal to the pooled rate (Table~\ref{tab:gen-x-config-hallucination}), and it has a failure type of its own, persona leakage: Gemini-3.1 writes the sampled persona's town and price preference into queries whose intent has neither (\S\ref{sec:judge-results}).

\paragraph{Diversity and alignment as proxies.}
Distinct-2 and Self-BLEU-4 flag a generator that recycles a few templates. Neither tracks parser F1. Alignment sKL does ($\rho=-0.917$, Table~\ref{tab:reference-alignment-matrix}), but only against the reference that produced it (\S\ref{sec:alignment}). Without labels, track sKL against a named reference (real traffic where available), watch synthetic $\kappa$ for generations bunching up, and hand-check a sample for missing constraints, unsupported additions and direction reversals.

\paragraph{Generator identity.}
Generator identity is a second source of risk, and the last column of Table~\ref{tab:gen-x-config-hallucination} shows the spread, from Qwen3's zero to more than half of the hard cases of Hunyuan-A13B and DeepSeek (\S\ref{sec:judge-results}). Hunyuan-A13B and DeepSeek also hold nearly all the direction reversals (\S\ref{sec:judge-results}). Adding syntax raises Gemini-3.1's share in all four matched pairs and Opus-4.6's in three, the two generators whose diversity also responds most to syntax. Hunyuan-A13B hallucinates in at least four of ten hard cases, whatever the prompt. These are generator-level patterns, descriptive only (\S\ref{sec:judge-results}, \S\ref{sec:limitations}). Ultimately, prompt recipes behave differently from model to model. Reasoning-capable LLMs writing short queries can still hallucinate.

\section{Conclusion}
\label{sec:conclusion}


We evaluated \nconfigs{} persona-conditioned prompt configurations across \nllms{} LLM generators at matched training volume, against \nhumaneval{} human-annotated and \nprod{} production queries. Conditioning pays. Every conditioned parser beats the vanilla prompt, and the IPIP personality bundle is the most reliable addition, $+1.1$ to $+2.1$ F1 in all four matched contrasts. Adjectives and syntax add smaller gains that partly overlap, and \texttt{pas}, \texttt{pa} and the mixed corpus form a statistically indistinguishable top group. Syntax alone moves generations towards real traffic in every contrast and drives diversity, at the plausible cost of the operator words where the audited defects concentrate. Personality widens the wording of each intent. Adjectives raise precision and leave recall almost unchanged. Alignment sKL is a good label-free proxy for parser F1 ($\rho=-0.917$ on the persona-card reference). For deployment we recommend the IPIP prompt as the default, adjectives when the parser over-extracts, and syntax when realism matters, with a check that operator words survive. Each new generator needs a hallucination audit of sampled generations.


\section{Limitations}
\label{sec:limitations}

\paragraph{Domain and distribution.}
The study is limited to one UK automotive marketplace and one search schema. Automotive search is a useful stress test because it contains many numeric, categorical, and operator-valued fields, but the findings are evidence about this domain and schema only.

\paragraph{UK-cultural exposure confound.}
The \texttt{u} condition assumes that generators can interpret UK geodemographic descriptors, postcode-area cues, and British automotive/search vocabulary. We do not know how much UK-specific data each LLM saw during pre-training, instruction tuning, localisation, or reinforcement learning. Since the generator pool includes models developed by organisations in different countries and deployment ecosystems, differences under \texttt{u} may reflect uneven exposure to British data rather than the demographic prompt design alone. We therefore avoid attributing behaviour to model nationality; the relevant unobserved factor is each model's training and alignment data composition.

\paragraph{Generator comparisons.}
Prompt comparisons are controlled within each generator, but generator-level patterns are descriptive. Different generators use different seed-intent slices, vendor stacks, decoding controls, and closed-model implementations. Per-generator diversity trends and audited-pool counts should therefore be read as diagnostics rather than as risk estimates or a definitive LLM leaderboard.

\paragraph{Conditional faithfulness audit.}
The committee audit is restricted to universal parser failures, so it estimates $P(\text{hallucination}\mid\text{universal parser failure})$ rather than full-corpus hallucination. Its three LLM judges overlap with generator families and are a triage instrument, not an oracle. The manual author-validation sample contains only $100$ cases from this same subset; one author cannot provide human--human agreement, per-configuration cells are too small for rate comparisons, and the sample does not estimate full-corpus label noise. Because pool entry requires every parser to fail, systematic drift that the parsers have absorbed is invisible to the audit.

\paragraph{Reasoning-model and scale restriction.}
The generator pool consists of relatively large reasoning-capable LLMs. Smaller models, non-reasoning chat models, or models run without explicit reasoning support may respond very differently to persona, syntax, and demographic prompts. They may show different diversity--faithfulness trade-offs, different rates of dropped or reversed constraints, and different sensitivity to UK-specific cues. Therefore, the prompt rankings reported here should not be applied directly to cheaper, smaller, or non-reasoning generators without re-running the same alignment, parser-utility, and faithfulness checks.

\paragraph{Prompt-axis sampling.}
The prompt families use different internal sampling schemes: IPIP trait levels, syntax modes, adjective classes, and UK demographic attributes are sampled from separate priors. The families also differ in token length, constraint wording, in-prompt demonstrations, and semantic guards, and we ran no length-matched placebo, so a bundle's effect is not separated from the extra instruction text it carries; \citet{kambhatla2025measuring} report that fine-grained persona detail can add little over a length control. The related finding of \citet{bui2025mixturepersonas} concerns mixing several complete personas within one generation and does not speak to combining the stylistic bundles used here, so it is not evidence for or against our composition choice. Alternative priors or bundle designs could shift the absolute diversity, alignment, and parser-utility scores, and our results should be read as evidence about these operationalisations rather than as an exhaustive search over prompt-sampling policies.

\section*{Ethics Statement}

\paragraph{Privacy and data minimisation.}
The marketplace data that seed generation are treated as anonymised structured search intents rather than user histories. No persistent user identifiers, session trajectories, or personal profiles are used in the experiments. The production reference of \S\ref{sec:production-reference} additionally uses real user \emph{query text}, which is not anonymous at source, so before any query left our infrastructure every logged string was screened for vehicle registration marks, full residential postcodes, telephone numbers, and first-person household detail, and matching rows were discarded rather than redacted ($257$ of $15{,}562$ rows, $1.65\%$, plus one blank query dropped before embedding). Discarding rather than masking avoids leaving partial identifiers in place, and structured browser and account context in the logs (postcode, latitude, longitude, session sort order) is dropped upstream by the schema map and never reaches the reference.

\paragraph{Human annotators.}
Human-authored queries are used only for evaluation and alignment analysis, not to infer annotator traits or train user-specific models. Annotator background information is reported only in aggregate.
Annotators were recruited through a University annotation campaign and rewarded with University credits.

\paragraph{Demographic prompting.}
The UK demographic prompts use synthetic area-level descriptors and must not be interpreted as modelling real individuals. Such prompts can encode stereotypes or proxy protected attributes. They should not be used for user profiling, eligibility decisions, personalised pricing, or demographic targeting.

\paragraph{Consumer-facing risk.}
Unfaithful synthetic queries can teach parsers to drop, add, or reverse constraints such as price limits, radius, finance terms, mileage, or sort order. In a live search product, these errors could affect which listings users see. Synthetic data should therefore be filtered or repaired before deployment, and user-facing systems should preserve confirmation, correction, or fallback mechanisms.

\section*{Acknowledgements}

We thank the annotators who took part in the University annotation campaign for their time and care. This work was funded by Autotrader UK, which also provided the data; we are grateful to our colleagues there for their support.

\bibliography{output_corrected}

\appendix
\renewcommand{\dbltopfraction}{0.8}\renewcommand{\topfraction}{0.9}\setcounter{dbltopnumber}{3}\setlength{\dbltextfloatsep}{14pt plus 2pt minus 4pt}\setlength{\textfloatsep}{14pt plus 2pt minus 4pt}

\section{Prompt Configuration Legend}
\label{app:prompts}
\begin{table}[ht]
\centering
\small
\setlength{\tabcolsep}{4pt}
\resizebox{\linewidth}{!}{%
\begin{tabular}{lccccl}
\toprule
\textbf{Code} & \textbf{UK} & \textbf{IPIP} & \textbf{Adj} & \textbf{Syn} & \textbf{Active conditioning} \\
\midrule
\texttt{n}   & --        & --        & --        & --        & no persona conditioning \\
\texttt{u}   & \checkmark & --        & --        & --        & UK demographic descriptors only \\
\texttt{p}   & --        & \checkmark & --        & --        & IPIP personality only \\
\texttt{a}   & --        & --        & \checkmark & --        & attribute-style adjectives only \\
\texttt{s}   & --        & --        & --        & \checkmark & syntactic mode only \\
\texttt{pa}  & --        & \checkmark & \checkmark & --        & IPIP + adjectives \\
\texttt{ps}  & --        & \checkmark & --        & \checkmark & IPIP + syntax \\
\texttt{as}  & --        & --        & \checkmark & \checkmark & adjectives + syntax \\
\texttt{pas} & --        & \checkmark & \checkmark & \checkmark & IPIP + adjectives + syntax \\
\bottomrule
\end{tabular}}
\caption{Legend for the \nconfigs{} prompt families. Columns indicate which of the four prompt axes are active: \textbf{UK} = UK demographic descriptors, \textbf{IPIP} = IPIP Big Five personality cues, \textbf{Adj} = attribute-style adjective injection, \textbf{Syn} = syntactic mode. \texttt{u} is the only family that uses the UK demographic axis; the other eight families combine subsets of the IPIP, adjective and syntax axes.}
\label{tab:prompt-families}
\end{table}

\section{IPIP Linguistic Mapping}
\label{app:ipip}

\begin{table}[ht]
\centering
\small
\begin{tabular}{p{0.30\linewidth}p{0.46\linewidth}}
\toprule
\textbf{Trait} & \textbf{Intended linguistic cues} \\
\midrule
Openness & exploratory wording, broader product framing, more descriptive modifiers \\
Conscientiousness & explicit constraint listing, precise budgeting, careful slot marking \\
Extraversion & energetic wording, higher affect, more direct engagement with the request \\
Agreeableness & polite requests, softened directives, cooperative framing \\
Neuroticism & hedging, uncertainty markers, reassurance-seeking language \\
\bottomrule
\end{tabular}
\caption{Operational mapping from IPIP personality axes to prompt-level linguistic cues.}
\label{tab:ipip-cues}
\end{table}

\paragraph{Persona pool and sampling.}
The neutral level carries $p_\text{neutral}=0.3$, giving up to $3^5=243$ profiles per generation call. 

\paragraph{Prompt assembly.}
For each non-neutral trait, the corresponding linguistic-cue snippet from Table~\ref{tab:ipip-cues} is concatenated into the system prompt; neutral traits contribute no text. This keeps the conditioning compact (at most five short cue strings per call).

\section{UK Demographic Persona Sampling}
\label{app:uk-persona}

The UK demographic prompt family \texttt{u} is anchored in a proprietary postcode-indexed attribute store assembled internally from UK Open Data sources (consumer-segmentation, census-derived, and lifestyle-survey aggregates), keyed by postcode area. The store is not publicly released, and the persona profiles are synthetic constructs (cf.\ Ethics Statement).

\paragraph{Pool size.}
The allocation covers 111 UK postcode areas, representing the most geographically distinct postcodes by visitor frequency. Each postcode is keyed to a county or city label (e.g.\ a Birmingham-area postcode resolves to \emph{Birmingham}); where the source assigns more than one county to a single postcode, the pipeline takes the first label for brevity in the prompt scaffold. Each postcode carries around 130 area-level attribute flags, drawn from a broader shared pool of around 560 distinct descriptions.

Each postcode's attribute list mixes coarse demographic indicators (age and occupation bands, household and housing context, financial-context bands) with lifestyle and consumer-style cues (shopping and payment habits, media and technology use, and attitudinal statements such as \emph{``It is worth paying extra for quality goods''}). The aggregates are banded rather than individual-level, which is what makes the resulting persona prompts read as plausible UK consumer voices rather than as identifying records.

\paragraph{Sampling.}
For each generation call assigned to the \texttt{u} family, the pipeline performs three steps:
\begin{enumerate}[noitemsep,topsep=2pt]
    \item Sample a postcode area uniformly from the 111 available.
    \item Sample $k \in [3, 15]$ attributes uniformly without replacement from that postcode's attribute list (default range; the lower bound of 3 is large enough to fill the prompt context given the cleaned descriptions).
    \item Inject the resulting bullet list into a system-prompt scaffold of the form \emph{``You are generating a search query as a person from \textbf{[location]} with the following characteristics: $\langle$bullets$\rangle$. Generate the query in a way that reflects this person's background, priorities, and typical concerns.''}
\end{enumerate}

\section{Syntactic Mode Sampling}
\label{app:syntax}

The syntactic axis (\S\ref{sec:prompt-types}) is implemented as a probabilistic choice over six modes calibrated against the web-query grammar of \citet{barr2008linguistic}, with probabilities normalised at sample time: noun phrase (head noun with modifiers, no verb) $0.40$, telegraphic (keywords only, stop-words removed) $0.22$, natural question (wh-words or auxiliary verbs) $0.18$, imperative (verb-initial command) $0.12$, post-modified noun phrase (noun with a prepositional phrase) $0.08$, and navigational (URL or site-name fragment) $0.001$.

\paragraph{Per-mode prompting.}
For each call assigned to a mode, a short instruction template specific to that mode is injected into the system prompt together with two examples sampled uniformly from a per-mode example bank (e.g.\ \emph{``Generate the query as a noun phrase without verbs or articles. Examples: $\langle$ex$_1\rangle$, $\langle$ex$_2\rangle$.''}). Each mode's example bank holds 10--20 hand-curated UK automotive queries; sampling fresh examples per call prevents the LLM from anchoring on a single canonical form.

\paragraph{Mode-wise faithfulness.}
Table~\ref{tab:syntax-hallucination} breaks the committee verdicts of \S\ref{sec:judge-results} down by the syntactic mode assigned to each generation.

\begin{table}[ht]
\centering
\small
\setlength{\tabcolsep}{4pt}
\begin{tabular}{lccc}
\toprule
\textbf{Mode} & $n$ & \textbf{Hallucinated} & \textbf{Rate} \\
\midrule
Noun phrase       & 228 & 99 & 0.434 \\
Telegraphic       & 134 & 56  & 0.418 \\
Imperative        & 32  & 11  & 0.344 \\
Natural question  & 57  & 21  & 0.368 \\
Post-modified NP  & 35  & 9   & \textbf{0.257} \\
\bottomrule
\end{tabular}
\caption{Committee positive-label share by syntactic mode within the selected universal-fail subset. Function-word-suppressing modes (noun phrase, telegraphic) carry the largest share; the post-modified noun phrase, which preserves relational prepositions, the smallest. These conditional shares are not mode-level risk estimates.}
\label{tab:syntax-hallucination}
\end{table}

\section{Attribute-Style Adjective Sampling}
\label{app:adjectives}

Table~\ref{tab:adjective-dimensions} gives the sampling probabilities and semantic gates for the eight dimensions of \S\ref{sec:prompt-types}.

\begin{table}[ht]
\centering
\small
\begin{tabular}{lcp{0.42\linewidth}}
\toprule
\textbf{Dimension} & $\Pr$ & \textbf{Semantic gate (illustrative)} \\
\midrule
\texttt{hard\_specs} & 0.20 & always eligible; no adjective injection \\
\texttt{economic}    & 0.18 & price field or \texttt{sort\_by=PRICE\_ASC} present \\
\texttt{affordance}  & 0.15 & body type / seat count compatible \\
\texttt{evaluative}  & 0.15 & mileage / condition / certification present \\
\texttt{vague}       & 0.12 & always eligible \\
\texttt{aesthetic}   & 0.08 & colour / body type present \\
\texttt{kinetic}     & 0.06 & engine / horsepower / transmission present \\
\texttt{social}      & 0.06 & luxury / exotic make only \\
\bottomrule
\end{tabular}
\caption{Configured probabilities and example semantic gates for the eight attribute-style adjective dimensions. Gates ensure adjective injection is consistent with the recoverable structured intent.}
\label{tab:adjective-dimensions}
\end{table}

\paragraph{Per-dimension prompting.}
For each call, five candidate adjectives are sampled without replacement from the chosen dimension's vocabulary (e.g.\ \emph{aggressive, sleek, sporty, elegant, matte} for \texttt{aesthetic}) and offered as optional suggestions (``Consider words like \dots''). For the six dimensions whose adjectives can imply product constraints, the prompt adds a semantic-fit guard: the adjective is used only when the provided search fields, the vehicle type or brand, and a natural shopper query support it, and omitted otherwise. \texttt{hard\_specs} injects no adjective and \texttt{vague} supplies low-commitment fillers (\emph{nice}, \emph{decent}, \emph{cracking}) treated as stylistic colour, so the axis varies style without adding intent constraints.


\section{Generator Pool Specification}
\label{app:generators}

For reproducibility, Table~\ref{tab:generator-pool} lists the exact model identifiers for the \nllms{} reasoning-mode LLMs in the generator pool (\S\ref{sec:prompt-types}).

Each generator runs under its provider's recommended inference regime, taking packaged configurations or production scripts first, then model-card ranges, then community defaults, with every parameter labelled by source. Closed reasoning APIs expose only documented reasoning effort, which is all we vary; for open-weight models we sample temperature, top-$p$, and top-$k$ within vendor safe bands, since decoding variation supports diversity--quality trade-offs \citep{zhang2024edt}.

\begin{table}[ht]
\centering
\footnotesize
\setlength{\tabcolsep}{3pt}
\begin{tabular}{@{}p{0.30\linewidth}p{0.22\linewidth}p{0.42\linewidth}@{}}
\toprule
\textbf{Model} & \textbf{Provider} & \textbf{Exact identifier} \\
\midrule
\multicolumn{3}{@{}l}{\emph{Closed-source frontier APIs}} \\
GPT-5.4 & OpenAI & \texttt{gpt-5.4} \\
Claude Opus 4.6 & Anthropic & \texttt{claude-\allowbreak opus-\allowbreak 4-6} \\
Gemini 3.1 Pro & Google DeepMind & \texttt{gemini-\allowbreak 3.1-\allowbreak pro-\allowbreak preview} \\
\midrule
\multicolumn{3}{@{}l}{\emph{Open-weight reasoning models (Hugging Face Hub)}} \\
DeepSeek-R1-\allowbreak Distill-Llama-70B & DeepSeek-AI & \texttt{deepseek-ai/\allowbreak DeepSeek-\allowbreak R1-\allowbreak Distill-\allowbreak Llama-\allowbreak 70B} \\
Hunyuan-\allowbreak A13B-\allowbreak Instruct & Tencent & \texttt{tencent/\allowbreak Hunyuan-\allowbreak A13B-\allowbreak Instruct} \\
Qwen3-Next-\allowbreak 80B-A3B-\allowbreak Thinking & Alibaba Qwen Team & \texttt{Qwen/\allowbreak Qwen3-\allowbreak Next-\allowbreak 80B-\allowbreak A3B-\allowbreak Thinking} \\
\bottomrule
\end{tabular}
\caption{Generator pool used to build the synthetic corpus. The top block lists closed-source frontier APIs; the bottom block lists open-weight reasoning models, with Hugging Face Hub identifiers shown verbatim.}
\label{tab:generator-pool}
\end{table}

\paragraph{Per-generator diversity.}
Table~\ref{tab:per-llm-selfbleu} gives the per-generator Self-BLEU-4 values behind the per-configuration means of Figure~\ref{fig:internal-diversity}. Lower values indicate greater diversity; the \textbf{Range} column shows how strongly each generator responds to prompt configuration. DeepSeek and GPT-5.4 are nearly flat across configurations, Hunyuan rises slightly, and Qwen3 moves the other way: every conditioned configuration makes its outputs more redundant than the vanilla prompt.

\begin{table*}[t]
\centering
\footnotesize
\setlength{\tabcolsep}{6pt}
\begin{tabular}{lccccccccc|c}
\toprule
\textbf{LLM} & \texttt{n} & \texttt{u} & \texttt{p} & \texttt{a} & \texttt{s} & \texttt{pa} & \texttt{ps} & \texttt{as} & \texttt{pas} & \textbf{Range} \\
\midrule
DeepSeek   & \cellcolor{green!10}0.606 & \cellcolor{green!15}0.580 & \cellcolor{green!15}0.580 & \cellcolor{green!15}0.582 & \cellcolor{green!14}0.587 & \cellcolor{green!15}0.584 & \cellcolor{green!15}0.582 & \cellcolor{green!14}0.586 & \cellcolor{green!14}0.585 & 0.026 \\
Gemini-3.1 & \cellcolor{green!13}0.591 & \cellcolor{green!8}0.620 & \cellcolor{green!10}0.610 & \cellcolor{green!17}0.572 & \cellcolor{green!42}0.440 & \cellcolor{green!10}0.608 & \cellcolor{green!54}0.376 & \cellcolor{green!47}0.409 & \cellcolor{green!55}0.368 & 0.251 \\
GPT-5.4    & 0.662 & 0.655 & 0.658 & 0.655 & 0.649 & 0.651 & 0.656 & 0.657 & 0.654 & 0.013 \\
Hunyuan    & \cellcolor{green!15}0.582 & \cellcolor{green!7}0.626 & \cellcolor{green!7}0.626 & \cellcolor{green!6}0.628 & \cellcolor{green!8}0.620 & \cellcolor{green!7}0.624 & \cellcolor{green!7}0.623 & \cellcolor{green!6}0.628 & \cellcolor{green!7}0.623 & 0.046 \\
Opus-4.6   & \cellcolor{green!15}0.582 & \cellcolor{green!32}0.489 & \cellcolor{green!21}0.547 & \cellcolor{green!23}0.537 & \cellcolor{green!36}0.469 & \cellcolor{green!26}0.523 & \cellcolor{green!42}0.436 & \cellcolor{green!41}0.440 & \cellcolor{green!45}0.422 & 0.160 \\
Qwen3      & \cellcolor{green!26}0.524 & \cellcolor{green!5}0.637 & 0.641 & 0.643 & 0.645 & 0.641 & 0.647 & 0.642 & 0.645 & 0.123 \\
\bottomrule
\end{tabular}
\caption{Self-BLEU-4 (lower is more diverse) per (LLM, configuration) cell on \texttt{split\_0}. Cell shading is on a global green scale (deeper green = more diverse); the \textbf{Range} column gives each LLM's max$-$min Self-BLEU across the nine configurations.}
\label{tab:per-llm-selfbleu}
\end{table*}

\section{Human Reference Data: Collection and Validation}
\label{app:rise-iaa}

This appendix gives the agreement metrics and filtering behind the \nhumaneval{}-pair corpus of \S\ref{sec:human-dataset}.

\paragraph{Validation and pooled IAA.}
Agreement metrics reflect both judge--annotator agreement (for persona-card items) and inter-annotator agreement (for website-front items). We observe a judge-anchored slot-value F1 of 0.85 for the persona-card protocol, while the website-front items show an inter-annotator slot-value F1 of 0.67, with disagreements stemming mostly from slot selection rather than value assignment.

\paragraph{Filtering pipeline.}
\label{sec:appendix:filtering}
Persona-card items are validated by a GPT-5.4 reasoning judge that compares each written NL query against the source structured intent; these items are human-written, so the judge's overlap with the generator pool does not bear on them. Items flagged as potentially inconsistent or as field-direction reversals (e.g.\ price floor vs.\ price ceiling) are manually reviewed by the lead researcher and dropped if the intent was not faithfully captured. Website-front: 176 items carry both annotation rounds. Pairs where either annotator's URL is missing (12 of 176) are excluded from the agreement metrics but not from the retained pool. Items where the two rounds disagree (132) are adjudicated by the lead researcher: 112 are resolved to one round and 20 where neither round is accepted are dropped, leaving 44 agreeing plus 112 adjudicated pairs, 156 in all.

\section{Corpus Comparison}
\label{app:corpus}

\begin{figure*}[t]
\centering
\includegraphics[width=\textwidth]{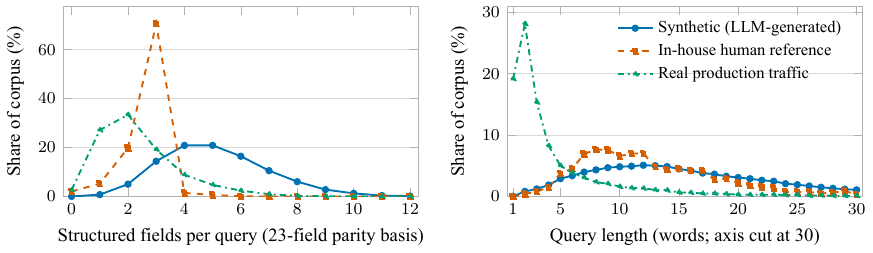}
\caption{Corpus distributions behind Table~\ref{tab:ecological-gap}. Left: share of each corpus by number of structured fields per query on the 23-field parity basis (axis cut at 12 fields). Right: share by query length in words, axis cut at 30 words, which covers 90\%, 93\% and 99\% of the synthetic, human and production corpora respectively. Production field counts as in Table~\ref{tab:ecological-gap}.}
\label{fig:corpus-distributions}
\end{figure*}

Table~\ref{tab:ecological-gap} places the three corpora on a common footing. Field counts are given on two bases because production gold is restricted to the $23$ parity-shared fields before it is written, so the $47$-field schema used elsewhere in the paper has no production counterpart; only the parity basis is like-for-like. The deployed engine is in a beta phase that parses these $23$ fields: make, model, variant, body type, colour, condition, drivetrain, fuel type and gearbox, bounds on mileage, year, engine size and price, and doors, seats, category, battery range, annual tax and search radius. The other $24$ fields of the schema are scheduled for the final release: finance and leasing terms (deposit, monthly price, contract term, annual mileage allowance), running costs (insurance group, CO$_2$ emissions, fuel consumption), electric-vehicle charging times, engine power, acceleration and boot size, listing filters (seller type, manufacturer-approved, Northern Ireland only, part exchange, write-off categories) and sort order. On that basis the human--production difference is $0.30$ fields. The length difference has no such explanation (\S\ref{sec:data-stats}): real traffic is dominated by short lookups, and its distinct-type rate of $0.705$ shows that the same strings recur, whereas the human reference is almost entirely unique. Figure~\ref{fig:corpus-distributions} plots the two distributions behind the table. Table~\ref{tab:field-category-guide} lists the schema by category.

\begin{table}[t]
\centering
\small
\setlength{\tabcolsep}{3.5pt}
\begin{tabular}{lccc}
\toprule
 & \textbf{Synth.} & \textbf{Human} & \textbf{Prod.} \\
\midrule
$n$                          & 109{,}098 & 1{,}214 & 15{,}304 \\
Fields, 47-basis (mean)      & 5.96 & 3.17 & --- \\
Fields, 23-basis (mean)      & 5.10 & 2.66 & 2.36 \\
Fields, 23-basis (median)    & 5 & 3 & 2 \\
Words (mean)                 & 17.64 & 14.31 & \textbf{4.85} \\
Words (median)               & 15 & 12 & \textbf{3} \\
Queries $\leq 2$ words       & 0.9\% & 0.4\% & \textbf{47.3\%} \\
Queries $\leq 4$ words       & 4.0\% & 2.7\% & \textbf{70.9\%} \\
Distinct-type rate           & 0.873 & 0.999 & 0.705 \\
\bottomrule
\end{tabular}
\caption{Corpus comparison. Synthetic counts are the nine per-configuration train+val pools used for the alignment fits; production field counts are computed on the $12{,}610$ rows where the deployed engine returned a parse, while length statistics use every retained row.}
\label{tab:ecological-gap}
\end{table}

\begin{table}[t]
\centering
\small
\setlength{\tabcolsep}{4pt}
\begin{tabular}{>{\raggedright\arraybackslash}p{0.31\linewidth}>{\raggedright\arraybackslash}p{0.60\linewidth}}
\toprule
\textbf{Category} & \textbf{Representative examples} \\
\midrule
Vehicle attributes & manufacturer, model, transmission \\
Vehicle specifications & mileage, registration year, engine size \\
Purchase price & minimum price, maximum price, total-price band \\
Finance/leasing & deposit, annual mileage allowance, contract length \\
Search controls & search radius, newest-first sorting, price sorting \\
Listing filters & manufacturer-approved listings, Northern Ireland only, part exchange available \\
Running costs & annual tax, insurance group, CO$_2$ emissions \\
Electric vehicles & battery range, charge time, quick-charge time \\
\bottomrule
\end{tabular}
\caption{The \nfields{}-field schema by category. Examples are representative schema constraints rather than exhaustive field lists.}
\label{tab:field-category-guide}
\end{table}

\section{Protocol- and Complexity-Stratified Robustness}
\label{app:robustness}
\begin{table}[!t]
\centering
\small
\setlength{\tabcolsep}{4pt}
\begin{tabular}{llrr}
\toprule
\textbf{Subset} & \textbf{Contrast} & \textbf{$\Delta$ F1} & \textbf{95\% CI} \\
\midrule
\multicolumn{4}{l}{\emph{By collection protocol}} \\
Persona-card & \texttt{n}$\rightarrow$\texttt{p}     & 2.019 & [1.425, 2.617] \\
Persona-card & \texttt{a}$\rightarrow$\texttt{pa}    & 1.568 & [1.071, 2.073] \\
Persona-card & \texttt{s}$\rightarrow$\texttt{ps}    & 2.082 & [1.562, 2.620] \\
Persona-card & \texttt{as}$\rightarrow$\texttt{pas}  & 1.204 & [0.740, 1.685] \\
Website-front & \texttt{n}$\rightarrow$\texttt{p}    & 1.931 & [0.271, 3.608] \\
Website-front & \texttt{a}$\rightarrow$\texttt{pa}   & 2.620 & [0.884, 4.441] \\
Website-front & \texttt{s}$\rightarrow$\texttt{ps}   & 2.517 & [0.994, 4.125] \\
Website-front & \texttt{as}$\rightarrow$\texttt{pas} & 0.230 & [$-1.464$, 1.724] \\
\midrule
\multicolumn{4}{l}{\emph{By gold-intent field count}} \\
0--2 fields & \texttt{n}$\rightarrow$\texttt{p}    & 2.018 & [0.493, 3.561] \\
0--2 fields & \texttt{a}$\rightarrow$\texttt{pa}   & 0.901 & [$-0.408$, 2.218] \\
0--2 fields & \texttt{s}$\rightarrow$\texttt{ps}   & 2.644 & [0.797, 4.537] \\
0--2 fields & \texttt{as}$\rightarrow$\texttt{pas} & 0.591 & [$-1.157$, 2.247] \\
3 fields & \texttt{n}$\rightarrow$\texttt{p}       & 2.328 & [1.515, 3.143] \\
3 fields & \texttt{a}$\rightarrow$\texttt{pa}      & 1.710 & [1.042, 2.395] \\
3 fields & \texttt{s}$\rightarrow$\texttt{ps}      & 2.042 & [1.434, 2.677] \\
3 fields & \texttt{as}$\rightarrow$\texttt{pas}    & 1.513 & [0.932, 2.116] \\
4+ fields & \texttt{n}$\rightarrow$\texttt{p}      & 1.554 & [0.672, 2.443] \\
4+ fields & \texttt{a}$\rightarrow$\texttt{pa}     & 2.018 & [1.191, 2.894] \\
4+ fields & \texttt{s}$\rightarrow$\texttt{ps}     & 2.069 & [1.278, 2.910] \\
4+ fields & \texttt{as}$\rightarrow$\texttt{pas}   & 0.664 & [$-0.021$, 1.406] \\
\bottomrule
\end{tabular}
\caption{Paired IPIP contrasts within subsets of the human reference: by collection protocol (persona-card $n=1{,}058$, website-front $n=156$) and by gold-intent field count ($n=174$ with $0$--$2$ fields, $608$ with $3$, $432$ with $4$+). The absolute protocol gap (persona-card F1 $79.6$--$83.2$ against website-front $51.7$--$55.2$ across the nine configurations) motivates reporting paired effects within, rather than pooling away, protocol. Overall-to-stratum rank agreement across the field-count strata is $\rho=0.933$, $0.983$ and $0.950$. All nine training configurations contain exactly the same $12{,}122$-intent multiset and field-count distribution, so the field-count rows test persistence across human intent complexity, not reweighted training.}
\label{tab:protocol-contrasts}
\end{table}

\begin{table}[!t]
\centering
\footnotesize
\setlength{\tabcolsep}{2pt}
\begin{tabular}{lrrr}
\toprule
\textbf{Human reference} & \textbf{$n$} & \textbf{$\rho$} & \textbf{95\% interval} \\
\midrule
Both protocols & 1{,}214 & $-0.800$ & [$-0.917$, $-0.633$] \\
Persona-card & 1{,}058 & $-0.917$ & [$-0.967$, $-0.783$] \\
Website-front & 156 & $-0.500$ & [$-0.800$, $-0.200$] \\
\bottomrule
\end{tabular}
\caption{Spearman association between vMF sKL (lower is closer) and F1 (higher is better) across nine prompt configurations. F1 is computed on each reference's own rows, whereas Table~\ref{tab:reference-alignment-matrix} ranks every column against the pooled F1 it displays, which is why the website-front $\rho$ is $-0.500$ here and $-0.417$ there. Intervals jointly bootstrap human rows $2{,}000$ times, recomputing both alignment and F1 ranks; the proportions below zero are 1.000, 1.000, and 0.999. This is an unlabelled alignment proxy, not an absolute-F1 predictor.}
\label{tab:alignment-bootstrap}
\end{table}

\paragraph{Parser training.}
All $\nconfigs{}+1$ Flan-T5-Large parsers were fully fine-tuned in bf16 on a single NVIDIA H200 for up to $10$ epochs with early stopping (patience $2$) and best-checkpoint selection on validation F1, at learning rate $2.5\times 10^{-4}$, effective batch $192$ without gradient accumulation, weight decay $0.01$, and warmup ratio $0.1$.

\paragraph{Bootstrap.}
Each row-level metric in Table~\ref{tab:protocol-contrasts} is first averaged across the three parser runs: for each configuration, we average the F1 of the three runs on every human test query. Bootstrap resampling is paired on human rows: we resample the \nhumaneval{} queries $100{,}000$ times and, in each resample, compare the two configurations of a contrast on the same queries.

\section{Surface Anatomy of the Prompt Axes}
\label{app:surface}
\input{arr_revision/surface_features.tex}

\input{arr_revision/axis_contrasts.tex}

This appendix holds the token-level measurements behind the mechanism reading in \S\ref{sec:discussion}. Table~\ref{tab:surface-features} measures what each axis does to the text itself, with the two reference corpora in the same units; Table~\ref{tab:axis-skl-contrasts} gives the matched-contrast alignment deltas with paired-bootstrap intervals, the alignment-side analogue of the F1 contrasts in \S\ref{sec:utility}. Function words, operator words and first-person forms are counted against fixed word lists: closed-class glue (articles, pronouns, auxiliaries, prepositions, conjunctions and contractions), $35$ operator words such as \emph{under}, \emph{within}, \emph{from} and \emph{newest}, and the first-person pronouns. With nine configurations, every cross-configuration correlation is descriptive.

\section{LLM Judge Protocol and Reliability}
\label{app:judge}

\begin{table*}[t]
\centering
\small
\setlength{\tabcolsep}{4pt}
\begin{tabular}{lrrrrrrr}
\toprule
\textbf{Comparator} & \textbf{$n$} & \textbf{Agreement [95\% CI]} & \textbf{$\kappa$} & \textbf{Precision} & \textbf{Recall} & \textbf{F1} & \textbf{TP/FP/FN/TN} \\
\midrule
Opus-4.7 & 100 & 91.0 [83.8, 95.2] & .775 & .767 & .920 & .836 & 23/7/2/68 \\
GPT-5.5 & 100 & 80.0 [71.1, 86.7] & .570 & .558 & .960 & .706 & 24/19/1/56 \\
Gemini-3.1 & 100 & 85.0 [76.7, 90.7] & .659 & .632 & .960 & .762 & 24/14/1/61 \\
Committee majority & 100 & 83.0 [74.5, 89.1] & .622 & .600 & .960 & .738 & 24/16/1/59 \\
\bottomrule
\end{tabular}
\caption{Author--committee agreement within the universal-fail subset after the author's targeted correction pass, under the rubric as run and the scoring rule of this appendix. Agreement is adjudicated rather than independent: the blinded first pass agreed on $72\%$. Hallucination is the positive class, and the final single-author label is the reference. Agreement intervals are 95\% Wilson intervals.}
\label{tab:manual-author-validation}
\end{table*}

\paragraph{Judges and aggregation.}
Each item is judged independently by \texttt{claude-opus-4-7} with adaptive thinking at maximum effort, \texttt{gpt-5.5-pro} with reasoning effort \texttt{xhigh}, and \texttt{gemini-3.1-pro-preview} with a high thinking level. The Opus and GPT judges are later releases than their generator counterparts, and the Gemini judge is the same model as the Gemini generator. Majority vote labels the $927$ universally failed rows, $345$ hallucinated and $582$ faithful. One scoring rule applies first: a $1{,}500$-mile radius is the marketplace's no-limit value, which generators and annotators alike leave unexpressed, so a verdict that faults only its omission does not count as a hallucination. All committee and author figures in the paper are reported after this rule.

\paragraph{Reliability.}
Across their valid verdicts, Opus-4.7 labels $23.4\%$ of cases hallucinated, GPT-5.5 $41.5\%$ and Gemini-3.1 $37.4\%$. Pairwise $\kappa$ is $0.871$ for GPT-5.5--Gemini-3.1, $0.632$ for Opus-4.7--Gemini-3.1 and $0.603$ for Opus-4.7--GPT-5.5, with Krippendorff's $\alpha=0.701$ across the three. GPT-5.5 and Gemini-3.1 are near-substitutable, and Opus-4.7 is more lenient.

\paragraph{Manual author audit.}
We draw a $100$-case random sample, allocated proportionally across the nine prompt configurations. The author sees only the structured intent and the generated query, and labels each case \emph{faithful} or \emph{hallucinated} under the judge prompt below. Prompt configuration, generator, parser outputs, committee labels and judge rationales stay hidden until all $100$ labels are submitted. A fully blinded first pass agreed with the committee majority on $72$ of $100$ cases. A second pass re-examined only the $16$ cases where a unanimous committee disagreed with the blinded verdict and corrected $11$ annotation slips (a missed constraint, an unnoticed reversal). The $83\%$ in Table~\ref{tab:manual-author-validation} is therefore adjudicated agreement, and all scores in the paper use the corrected labels. Sixteen of the $17$ remaining disagreements are cases the author labels faithful and the committee labels hallucinated.

\paragraph{Rationale cues.}
The four patterns of \S\ref{sec:judge-results} are tagged on the rationale text with hand-audited, word-boundary cue lists: \textbf{dropped constraints} (missing, omits, drops, absent, not in, not encoded, not expressed, not reflected, not present, not mentioned, not included, never mentions), \textbf{unsupported additions} (introduces, adds, extra, additional, not in the intent, not present in the intent), \textbf{direction reversals} (reverses, contradicts, opposite, instead of, flips) and \textbf{swapped values} (substitutes, replaces, wrong value, different value, changes the value, mismatched).

\newpage
\paragraph{Judge prompt.}
The judge system prompt is:

{\scriptsize
\begin{verbatim}
You are evaluating whether a natural-language
search query faithfully encodes a given set of
intent constraints. You will be shown:

  INTENT: a list of (field, value) constraints
    the query was supposed to express.
  QUERY: the natural-language query produced
    from that intent.

Decide: did the QUERY faithfully encode the
INTENT, or did it hallucinate?

FAITHFUL = every INTENT constraint is
recoverable from a reasonable reading of the
QUERY (paraphrase, synonym, indirect
formulation, unusual word order all count as
faithful) AND the QUERY does not introduce any
constraint not in the INTENT.

A fuzzy or approximate rendering of a
constraint is acceptable as long as it is a
sensible match for the original value. For
example:
  - INTENT search_max_mileage: 40000
      -> QUERY "low mileage"   = FAITHFUL
  - INTENT search_make: ["Honda"]
      -> QUERY "Japanese car"  = FAITHFUL
  - INTENT search_manufactured_year_from: 2018
      -> QUERY "around 2020"   = HALLUCINATED
      (changes which years pass the constraint)
  - INTENT search_make: ["Honda"]
      -> QUERY "Toyota"        = HALLUCINATED

HALLUCINATED = at least one of:
  - the QUERY adds a constraint not in the
    INTENT;
  - the QUERY drops a constraint from the
    INTENT;
  - the QUERY reverses or contradicts a
    constraint (e.g. "over X" when the INTENT
    said "under X");
  - the QUERY substitutes a value that is not
    a sensible fuzzy match for the INTENT
    value.

Rare phrasing alone is not hallucination. A
fuzzy term that sensibly maps to the intent
value is not hallucination.

Return JSON only, with exactly two keys:
  {"hallucinated": true | false,
   "rationale": "<one short sentence>"}

Do not include any text outside the JSON
object.
\end{verbatim}
}

\end{document}

%% file: arr_revision/reference_alignment_matrix.tex
\begin{table*}[t]
\centering
\small
\setlength{\tabcolsep}{6pt}
\begin{tabular}{lllcccc}
\toprule
 & & & \multicolumn{3}{c}{\textbf{In-house human reference}} & \multicolumn{1}{c}{\textbf{Real users}} \\
\cmidrule(lr){4-6}\cmidrule(lr){7-7}
\textbf{Cfg} & \textbf{F1}$\uparrow$ & \textbf{Spread $\kappa$}$\downarrow$ & \textbf{Persona-card} & \textbf{Website-front} & \textbf{Both protocols} & \textbf{Production} \\
 & & & $n{=}1{,}058$ & $n{=}156$ & $n{=}1{,}214$ & $n{=}15{,}304$ \\
\midrule
\texttt{pas} & \cellcolor{teal!42}$79.62_{\text{1st}}$ & \textbf{3641} & \cellcolor{teal!42}$\mathbf{264.6}_{\text{1st}}$ & \cellcolor{teal!42}$\mathbf{438.5}_{\text{1st}}$ & \cellcolor{teal!42}$\mathbf{252.5}_{\text{1st}}$ & \cellcolor{teal!42}$\mathbf{812.5}_{\text{1st}}$ \\
\texttt{pa} & \cellcolor{teal!39}$79.40_{\text{2nd}}$ & 3786 & \cellcolor{teal!31}$279.5_{\text{4th}}$ & \cellcolor{teal!13}$497.5_{\text{7th}}$ & \cellcolor{teal!27}$273.5_{\text{5th}}$ & \cellcolor{teal!5}$928.7_{\text{8th}}$ \\
\texttt{u} & \cellcolor{teal!37}$79.19_{\text{3rd}}$ & 3756 & \cellcolor{teal!33}$277.0_{\text{3rd}}$ & \cellcolor{teal!20}$483.7_{\text{5th}}$ & \cellcolor{teal!30}$269.5_{\text{3rd}}$ & \cellcolor{teal!16}$894.1_{\text{5th}}$ \\
\texttt{ps} & \cellcolor{teal!35}$79.05_{\text{4th}}$ & 3653 & \cellcolor{teal!38}$270.1_{\text{2nd}}$ & \cellcolor{teal!38}$447.2_{\text{2nd}}$ & \cellcolor{teal!38}$258.5_{\text{2nd}}$ & \cellcolor{teal!39}$821.6_{\text{2nd}}$ \\
\texttt{as} & \cellcolor{teal!30}$78.55_{\text{5th}}$ & 3681 & \cellcolor{teal!29}$281.9_{\text{5th}}$ & \cellcolor{teal!34}$455.5_{\text{3rd}}$ & \cellcolor{teal!30}$269.9_{\text{4th}}$ & \cellcolor{teal!36}$830.9_{\text{3rd}}$ \\
\texttt{p} & \cellcolor{teal!23}$77.99_{\text{6th}}$ & 3789 & \cellcolor{teal!29}$282.4_{\text{6th}}$ & \cellcolor{teal!12}$500.0_{\text{8th}}$ & \cellcolor{teal!25}$276.4_{\text{7th}}$ & \cellcolor{teal!6}$927.1_{\text{7th}}$ \\
\texttt{a} & \cellcolor{teal!20}$77.70_{\text{7th}}$ & 3838 & \cellcolor{teal!2}$318.7_{\text{8th}}$ & $524.8_{\text{9th}}$ & $311.1_{\text{9th}}$ & $945.0_{\text{9th}}$ \\
\texttt{s} & \cellcolor{teal!11}$76.91_{\text{8th}}$ & 3691 & \cellcolor{teal!25}$287.2_{\text{7th}}$ & \cellcolor{teal!29}$465.7_{\text{4th}}$ & \cellcolor{teal!25}$275.8_{\text{6th}}$ & \cellcolor{teal!33}$841.5_{\text{4th}}$ \\
\texttt{n} & $75.98_{\text{9th}}$ & 3869 & $321.4_{\text{9th}}$ & \cellcolor{teal!14}$495.6_{\text{6th}}$ & \cellcolor{teal!1}$309.8_{\text{8th}}$ & \cellcolor{teal!14}$902.2_{\text{6th}}$ \\
\midrule
\multicolumn{3}{l}{Reference spread $\kappa$} & 3323 & 3116 & 3233 & 2407 \\
\multicolumn{3}{l}{Spearman $\rho$ (rank vs.\ F1)} & $-0.917$ & $-0.417$ & $-0.800$ & $-0.350$ \\
\bottomrule
\end{tabular}
\caption{Alignment sKL (von Mises--Fisher symmetric KL, lower is closer) from each prompt configuration to each reference, with every value's within-column rank as a subscript and cell shading on each column separately (darker is better: higher F1, lower sKL). \textbf{Spread $\kappa$} is the synthetic cloud's own concentration (lower is broader). Rows are ordered by tuple F1 on the in-house test set, so a proxy that tracked utility perfectly would read $1\text{st}, 2\text{nd}, \dots$ down each column; F1 is identical across columns because no parser was scored on production text. Absolute sKL is not comparable across columns (real traffic is more dispersed); only the within-column ranking is. All three production views give identical rankings ($\tau{=}1.000$).}
\label{tab:reference-alignment-matrix}
\end{table*}

%% file: arr_revision/surface_features.tex
\begin{table}[t]
\centering
\small
\setlength{\tabcolsep}{3.5pt}
\resizebox{\linewidth}{!}{%
\begin{tabular}{lcccccc}
\toprule
\textbf{Cfg} & \textbf{Words} & \textbf{Func.\%} & \textbf{1st-p.\%} & \textbf{Oper.\%} & \textbf{Types} & \textbf{Cov.\%} \\
\midrule
\texttt{n} & 15.26 & 29.7 & 10.6 & 92.1 & 1{,}799 & 60.9 \\
\texttt{u} & 18.90 & 33.1 & 17.4 & 92.3 & 2{,}615 & 73.6 \\
\texttt{p} & 20.84 & 34.1 & 19.0 & 92.3 & 2{,}582 & 69.6 \\
\texttt{a} & 16.50 & 29.4 & 10.8 & 92.2 & 1{,}801 & 63.6 \\
\texttt{s} & 15.74 & 28.8 & 10.9 & 88.4 & 1{,}858 & 61.4 \\
\texttt{pa} & 21.47 & 34.4 & 20.0 & 92.4 & 2{,}649 & 71.0 \\
\texttt{ps} & 17.10 & 30.1 & 16.3 & 88.2 & 2{,}407 & 68.8 \\
\texttt{as} & 15.72 & 28.7 & 11.1 & 88.2 & 1{,}839 & 63.6 \\
\texttt{pas} & 17.25 & 29.9 & 16.1 & 87.6 & 2{,}612 & 70.4 \\
\midrule
\multicolumn{7}{l}{\emph{Axis effect (mean over the four matched contrasts)}} \\
$+$\texttt{p} & +3.36 & +2.92 & +7.01 & -0.11 & +738 & +7.6 \\
$+$\texttt{a} & +0.50 & -0.06 & +0.30 & -0.13 & +64 & +2.0 \\
$+$\texttt{s} & -2.07 & -2.54 & -1.48 & -4.18 & -29 & -0.2 \\
\midrule
In-house human reference & 14.31 & 42.3 & 58.7 & 74.5 & --- & 51.8 \\
Real production traffic & 4.85 & 20.8 & 5.2 & 16.9 & --- & --- \\
\bottomrule
\end{tabular}}
\caption{Surface anatomy of the prompt axes. \textbf{Words}: mean query length (whitespace tokens, as in Table~\ref{tab:ecological-gap}). \textbf{Func.\%}: function-word share of tokens, from a fixed closed-class list released with the code. \textbf{1st-p.\%}: share of queries containing a first-person form. \textbf{Oper.\%}: share of queries containing an operator-bearing word (\emph{under}, \emph{within}, \emph{from}, \emph{newest}, \dots); the operator list overlaps the function-word list, so the two columns are related signals, not independent ones. \textbf{Types}: distinct word types in a fixed $100$k-token sample (seed $20260801$) --- vocabulary breadth at matched budget; the nine banks are row-matched rather than token-matched, and the anchor corpora are too small for the budget, so their value is withheld. \textbf{Cov.\%}: share of the $1{,}000$ most frequent production content words (glue and bare numbers excluded) present in the corpus vocabulary --- lexical coverage, not distributional match. Adding \texttt{s} deletes grammatical glue, operator words included (the realism direction); adding \texttt{p} adds glue of the annotator-register kind (pronouns, politeness) while leaving operator presence untouched, and widens the vocabulary at fixed intent; adding \texttt{a} moves no surface channel. Real traffic is nearly glue- and first-person-free, while the in-house reference is majority first-person. All values are descriptive statistics over nine configurations.}
\label{tab:surface-features}
\end{table}

%% file: arr_revision/axis_contrasts.tex
\begin{table}[t]
\centering
\small
\setlength{\tabcolsep}{3pt}
\resizebox{\linewidth}{!}{%
\begin{tabular}{lcccc}
\toprule
 & \multicolumn{2}{c}{\textbf{Real traffic}} & \multicolumn{2}{c}{\textbf{In-house, both protocols}} \\
\cmidrule(lr){2-3}\cmidrule(lr){4-5}
\textbf{Contrast} & $\Delta$ sKL & \textbf{95\% CI} & $\Delta$ sKL & \textbf{95\% CI} \\
\midrule
\multicolumn{5}{l}{\emph{Adding \texttt{s}}} \\
\texttt{n}$\rightarrow$\texttt{s} & $-60.7$ & $[-61.4, -60.0]$ & $-33.9$ & $[-35.9, -31.9]$ \\
\texttt{p}$\rightarrow$\texttt{ps} & $-105.5$ & $[-106.1, -104.9]$ & $-17.9$ & $[-21.1, -14.7]$ \\
\texttt{a}$\rightarrow$\texttt{as} & $-114.1$ & $[-114.7, -113.6]$ & $-41.2$ & $[-43.9, -38.5]$ \\
\texttt{pa}$\rightarrow$\texttt{pas} & $-116.1$ & $[-116.8, -115.4]$ & $-21.0$ & $[-24.5, -17.5]$ \\
\midrule
\multicolumn{5}{l}{\emph{Adding \texttt{p}}} \\
\texttt{n}$\rightarrow$\texttt{p} & $+24.9$ & $[+24.2, +25.6]$ & $-33.3$ & $[-36.5, -30.1]$ \\
\texttt{a}$\rightarrow$\texttt{pa} & $-16.4$ & $[-16.8, -16.0]$ & $-37.6$ & $[-40.2, -35.1]$ \\
\texttt{s}$\rightarrow$\texttt{ps} & $-19.9$ & $[-20.1, -19.8]$ & $-17.3$ & $[-18.3, -16.4]$ \\
\texttt{as}$\rightarrow$\texttt{pas} & $-18.4$ & $[-18.6, -18.2]$ & $-17.4$ & $[-18.4, -16.3]$ \\
\midrule
\multicolumn{5}{l}{\emph{Adding \texttt{a}}} \\
\texttt{n}$\rightarrow$\texttt{a} & $+42.9$ & $[+42.4, +43.3]$ & $+1.3$ & $[-0.8, +3.3]$ \\
\texttt{p}$\rightarrow$\texttt{pa} & $+1.6$ & $[+1.5, +1.7]$ & $-2.9$ & $[-3.4, -2.5]$ \\
\texttt{s}$\rightarrow$\texttt{as} & $-10.6$ & $[-10.7, -10.5]$ & $-5.9$ & $[-6.4, -5.5]$ \\
\texttt{ps}$\rightarrow$\texttt{pas} & $-9.0$ & $[-9.1, -9.0]$ & $-5.9$ & $[-6.3, -5.6]$ \\
\bottomrule
\end{tabular}}
\caption{Matched-contrast change in Alignment sKL (negative is closer) from adding each prompt axis, against real production traffic and the in-house reference (both protocols). Intervals are $2{,}000$-replicate paired bootstraps over reference rows (seed $20260801$): each replicate resamples the reference once and evaluates all configurations against it, so the four contrasts of an axis are paired; synthetic fits are held at their full-corpus values, as in every other interval reported for the alignment analysis. Adding \texttt{s} moves generations toward real traffic in all four contrasts; adding \texttt{p} moves them toward the in-house reference in all four but not toward real traffic (\texttt{n}$\rightarrow$\texttt{p} is positive), and \texttt{a} helps only when \texttt{s} is present.}
\label{tab:axis-skl-contrasts}
\end{table}